\pdfoutput=1
\documentclass[11pt]{article}

\usepackage[final]{acl}
\usepackage{amsmath}

\usepackage{times}
\usepackage{latexsym}

\usepackage[T1]{fontenc}
\usepackage[utf8]{inputenc}

\usepackage{microtype}

\usepackage{inconsolata}

\usepackage{graphicx}
\usepackage{xcolor}
\usepackage{todonotes}
\usepackage{xurl}

\definecolor{lightblue}{RGB}{154, 196, 246}
\definecolor{lightorange}{RGB}{247, 225, 153}
\definecolor{lightpurple}{RGB}{206, 120, 251}
\definecolor{lightgreen}{RGB}{170, 255, 200}

\DeclareSymbolFont{extraup}{U}{zavm}{m}{n}
\DeclareMathSymbol{\varheart}{\mathalpha}{extraup}{86}
\DeclareMathSymbol{\vardiamond}{\mathalpha}{extraup}{87}
\DeclareMathSymbol{\varclub}{\mathalpha}{extraup}{88}

\newcommand{\mysubsection}[1]{\vspace{0.3em}\noindent\textbf{#1}}
\newcommand{\systemname}[0]{\textsc{Isca}}

\title{\systemname{}: A Framework for \underline{I}nterview-\underline{S}tyle \underline{C}onversational \underline{A}gents}

\author{Charles Welch$^\heartsuit$ \and Allison Lahnala$^\heartsuit$ \and Vasudha Varadarajan$^\varheart$ \and \\ \textbf{Lucie Flek}$^\diamondsuit$ \and \textbf{Rada Mihalcea}$^\vardiamond$ \and \textbf{J. Lomax Boyd}$^\clubsuit$ \and \textbf{João Sedoc}$^\spadesuit$\\ 
    $^\heartsuit$McMaster University, 
    $^\varheart$Stony Brook University,
    $^\diamondsuit$University of Bonn,\\ $^\vardiamond$University of Michigan, $^\clubsuit$Johns Hopkins University, $^\spadesuit$New York University \\
    \texttt{cwelch@mcmaster.ca}
}

\begin{document}
\maketitle
\begin{abstract}
We present a low-compute non-generative system for implementing interview-style conversational agents which can be used to facilitate qualitative data collection through controlled interactions and quantitative analysis. Use cases include applications to tracking attitude formation or behavior change, where control or standardization over the conversational flow is desired. We show how our system can be easily adjusted through an online administrative panel to create new interviews, making the tool accessible without coding. Two case studies are presented as example applications, one regarding the Expressive Interviewing system for COVID-19 and the other a semi-structured interview to survey public opinion on emerging neurotechnology. Our code is open-source, allowing others to build off of our work and develop extensions for additional functionality.
\end{abstract}

\section{Introduction}

Conversational agents are increasingly used for applications in healthcare~\cite{valizadeh-parde-2022-ai}, customer service and engagement~\cite{soni2023large}, and education~\cite{yan2024practical}. They can help motivate behavior change, assisting with addiction~\cite{he2022can}, medication adherence, and healthy behaviors~\cite{aggarwal2023artificial}.  They are also useful tools for understanding thoughts, opinions, and concerns~\cite{ziems2024can}. 

In this paper, we describe our framework \systemname{}, for interview style conversational agents for applications to health, behavior change, and information gathering. \systemname{} enables researchers to implement standardized interviews with customizable conversation flows. It is deliberately non-generative, allowing for more controlled interactions across participants and avoiding off-topic discussion. The backend of the system includes language detection modules, which the researcher can customize through the admin portal to design rules for triggering follow-up questions based on elements of the user's responses. The system is equipped to deploy the studies within a participant-facing interface. Language analytics are built-in, both for the researcher's admin view to gain insight into the study population's responses and for the participants to gain insight into their own language.  Our system is run as a web server built on the Django framework~\cite{django} and allows users to easily configure interviews without coding. 

\systemname{} can be used to understand attitude formation, attitude polling, and monitoring behavior change, using multiple choice survey questions given before and after each interview.
We present two case studies: 1) to address stress related to the COVID-19 pandemic~\cite{welch-etal-2020-expressive} and 2) to survey public opinion about human brain organoids, an emerging neurotechnology that raises ethical concerns among segments of the public. Finally, we discuss applications and extensions. We intentionally avoid large language models and generative approaches as they are unnecessary and costly. Our code\footnote{\url{https://github.com/cfwelch/framework-interview-style-agents}} and a demo video\footnote{\url{https://youtu.be/_5XvMsZf8dA}} are made publicly available, so that others can easily set up and run servers locally or extend functionality.

\begin{figure}[t]
    \centering
    \includegraphics[width=\linewidth]{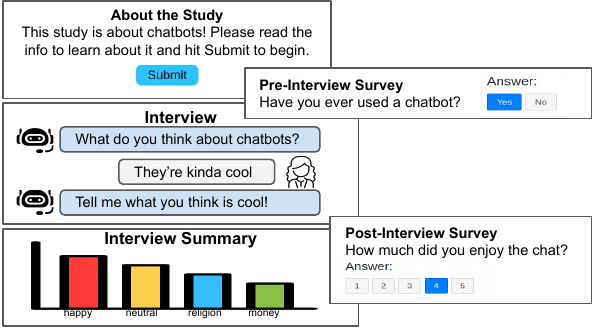}
    \caption{\systemname{}: An Interview-Style Framework}
    \label{fig:framework_illustration}
\end{figure}

\section{Related Work}

\mysubsection{Surveys and Interviews} 
Recent research has explored advancements in language-based survey methodologies, particularly focusing on \textit{tailoring} approaches to enhance response rates and reduce respondent burden~\cite{dillman2014internet, sikstrom2023precise}.
Computerized adaptive testing (CAT) has long been recognized for its effectiveness in selecting informative questions based on participants' previous responses. Recently, it has emerged as a viable method to traditional fixed-format surveys~\cite{wainer2000computerized, varadarajan-etal-2024-alba}. While CAT effectively improves engagement and response rates, it faces several limitations including the need for extensive pre-calibration with large but fixed question banks and significant computational resources. Recent solutions employ humans-in-the-loop to aid with question selection to train reinforcement learning models to dynamically update survey questions~\cite{velez2024crowdsourced}.

Designing interviews with meaningful outcomes requires \textit{careful} consideration of the context, role dynamics, trust and expectations between the involved parties~\cite{schilling2013surveys}. Distinct interviewing styles can elicit distinct response patterns from the participants reflective of their cognitive patterns~\cite{priede2011comparing}.
However, with the advent of conversational agents and chatbot-based interviewing systems, the elicited responses can differ due to the personality of the agent~\cite{xiao2020tell}, anthropomorphism~\cite{rhim2022application} or the mode of interaction~\cite{oates2022audio, zarouali2024comparing}. AI-powered chatbots have been explored as a means to provide telehealth advice~\cite{xu2024talk2care}, mental health assessments~\cite{schick2022validity} and facilitating discussions~\cite{nguyen2023role}. AI chatbots present challenges including data privacy concerns, ethical considerations in data handling, and potential impacts on authentic human interactions~\cite{adam2021ai, marks2023ai}.

\mysubsection{Conversational Agent Design}
Several platforms and frameworks exist for developing conversational agents. 
Rasa, available in both open-source and commercial versions, specializes in machine learning-based dialog management using slot-value systems~\cite{bocklisch2017rasa}. 
ParlAI provides a research platform specifically designed for dialog research~\cite{miller-etal-2017-parlai}, while AIML and ChatScript offer rule-based chatbot development through specialized scripting languages~\cite{ramesh2017survey}.
Similar to \systemname{}, Juji offers conversational capabilities~\cite{juji}, while Riff represents a more focused application - a generative chatbot system designed to enhance college students' learning through reflection~\cite{cavagnaro2023riff}, specifically targeting users 18 and older to avoid concerns associated with generative AI.
Researchers used  Juji for interactive surveying~\citep{xiao2020tell}, however, this is closed-source limiting the ability to tailor their infrastructure. 

\section{System Overview}\label{sec:system-overview}

\begin{figure}[t]
    \centering
    \includegraphics[width=\linewidth]{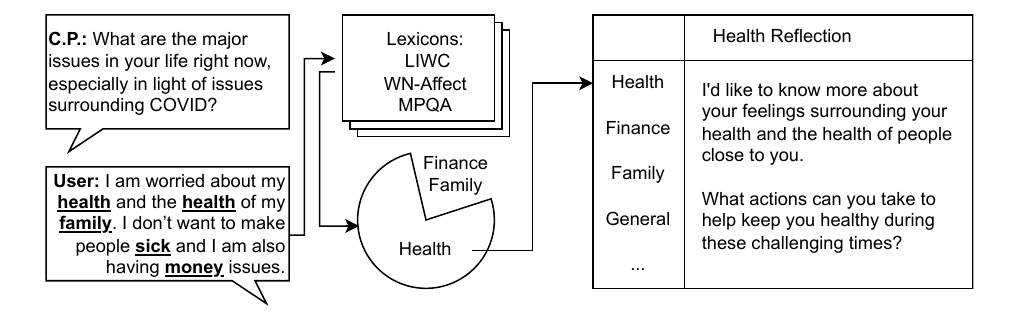}
    \caption{Main questions and reflection triggers.}
    \label{fig:questions_reflections}
\end{figure}

We release \systemname{}, an open framework for researchers to create conversational interview-style social science studies. As illustrated in Figure~\ref{fig:framework_illustration}, \systemname{} facilitates the implementation of the interview flow and surveys that can be administered before and after the interview. It offers a unique experience for participants by providing analytical summaries and visualizations of what they wrote.

Our conversational interview design revolves around \textit{main questions} and \textit{reflections}. The main questions will be asked to every participant by completion of the interview, but participants may encounter them at different times depending on the conversation flow. The conversation flow is impacted by the content of the participant's turns, as the language triggers specific reflections (i.e., follow-up questions) when certain trigger conditions are satisfied. Figure~\ref{fig:questions_reflections} shows an example where a participant's response with a high proportion of \textit{health} topic words triggers a \textit{health} reflection, prompting the user to reflect more on health.

\mysubsection{Participant Flow.} Participants can be directed to the website where the project is hosted either to a page that shows all topics with active interviews, or directly linked to the start page for a specific topic. The start page shows an informational screen about the purpose of the interview. This can serve as a disclosure of information about the present study and asking for consent. Next, they are sent to the first multiple choice page. This page can be used for checking pre-interview participant status by asking yes/no or Likert scale questions. After answering questions, they are directed to the conversation.

The conversation begins with an introduction from the chatbot and an initial main question. When the participant responds, a reflection may trigger. One reflection can be triggered in between each main question. If no defined reflections trigger, preprogrammed generic reflections (e.g. ``Tell me a little more about that'') can trigger instead. These will trigger if the participant has taken less than 15 seconds to write their response or if they type less than 100 characters. Only one of these type of triggers can occur in a conversation, as we found that repeatedly reminding someone to write more tends to be frustrating and causes them to disengage. After all the main questions from the active interview are asked, the conversation will end. The participant is then redirected to post-interview multiple choice questions. After they respond to these questions, they are shown a summary of their interaction (Figure \ref{fig:visual-summary} in the Appendix shows most discussed categories) and have options to download their data, reset the page if they want to start the conversation over, or provide feedback.

\section{Administrative Panel Demo}
\label{sec:administrative-panel}

The administrative panel consists of seven main pages with distinct functions. Admins can create interview topics, define interview questions for a topic, manage lexicon categories, configure survey questions, view analytics on the dashboard, perform topic modeling, and manage the FAQ. While pages like the lexicon management are globally managed, other pages, such as the interviews, are specific to a topic. On these pages, the dropdown in the sidebar can be changed at any time to reload the page with information for that topic. Figures showing each of the pages are in Appendix~\ref{sec:appendix-interface} and step-by-step documentation is provided in the supplementary material.

\mysubsection{Topics.} The topics page shows the list of conversation topics (Figure \ref{fig:visual-topic-page}). Each topic can have one active interview at a time. When adding a new topic, admins can provide a name and icon for the topic, the name of the chatbot, and the intro/disclaimer screen that shows before any conversation starts.

\mysubsection{Interviews.} The interview overview page (Figure \ref{fig:visual-interview-overview}) shows the list of interviews for a topic. A new interview can be created when one wants to update the questions that will be asked in an interaction. Any notes on the topic will be shown on this page. When adding a new interview, admins can add questions and reflections. 
Admins can choose to trigger the reflection when there is a dominant category from the set of lexicons that are active for that topic, sentiment, and whether or not a different reflection has already been triggered. A dominant category is one that occurs more than 50\% more often in the response than the next most frequent lexical category. The sentiment option uses VADER to efficiently classify each utterance as positive, negative, or neutral~\cite{hutto2014vader}. 
On the overview page, clicking on any of these interviews shows details and allows setting the active interview (Figure \ref{fig:visual-interview-details}). 

\mysubsection{Lexicons.} The lexicons page (Figure \ref{fig:visual-lexicons}) allows admins to define new lexicon categories and to add/remove words from each category. When adding words, admins provide a comma-separated list of words and word stems. Word stems end with an asterisk and will match words beginning with the preceding stem, while other words must exactly match (case insensitive). When the user visits the lexicon topic management page (Figure \ref{fig:visual-lexicon-topics}), they can assign lexicon categories to a topic or remove them. The categories active for each topic are listed on this page. An active category means that those categories can be used as reflection triggers when adding an interview, and that those categories will be detected and recorded for summary statistics and for the administrative dashboard.

\mysubsection{Surveys.} The survey management page (Figure \ref{fig:visual-survey}) shows the list of questions for a topic. Questions can be added or deleted. The intro and outro boxes allow admins to toggle if the question is asked before or after the interview (or both). Each question can be a yes/no question or a Likert scale question. The responses to questions defined here will be visualized on the administrative dashboard.

\mysubsection{FAQs.} Admins can add or delete frequently asked questions in the FAQ page (Figure \ref{fig:visual-faqs}).  This page is accessible to participants who are currently having a conversation either through the side navigation bar or through the conversational interface directly when they ask the bot a question. When a question is detected, a notification will appear that states that the interview was not designed for users to ask the bot questions and instead provides a link to the FAQ page for that topic.

\mysubsection{Dashboard.} The dashboard shows a variety of aggregate statistics over interviews for the selected topic. Figure \ref{fig:visual-dashboard-top} shows the top of the dashboard, which contains a bar chart showing the most frequently discussed topics. Charts on this page use lexicon categories that are active for the topic. We also see the total number of conversations, average response length and interview time. The interview time and response length statistics can be selected, which redirects the admin to distribution plots for each statistic.
The next plot on the dashboard shows the detected word categories (Figure \ref{fig:visual-dashboard-word-categories}). Unlike the previous plot, which shows the total number of conversations a given category appears in, this plot shows the frequency distribution of each category across conversations.
The dashboard also shows a list of summaries and plots. Each summary has a date, word count, and button that redirects to the participant-facing summary for that page.
Each survey question is used to generate a plot. Yes/no questions generate bar plots and Likert scale questions generate line plot distributions (Figure \ref{fig:visual-dashboard-likert}) with an entry for before and/or after the interview.

\mysubsection{Topic Modeling.} From the dashboard, you can go to separate topic modeling pages for LDA~\cite{blei2003latent} and BERTopic~\cite{grootendorst2022bertopic}. The pages are identical except each allows you to run a different topic modeling method, which can be used for exploration of interview responses. The topic modeling pages show an overview of the previous topic modeling runs (Figure \ref{fig:topic-overview}). The admin can choose a number of topics and start a new topic model, which will run as a subprocess. The status of the previous runs is shown below and will update automatically, as it regularly checks the subprocess status. The table of previous runs shows the date, number of topics, how long it took to run, and the topic coherence.

Clicking to see the results of the previous topic modeling runs takes the admin to a page specifically for those results (Figure \ref{fig:topic-visual-page}). This page shows the distribution of each topic's frequency and top ten words. Clicking on these topics shows a list of all conversation turns where the words occurred for more context. The bottom of this page contains the visualization generated with pyLDAvis~\cite{sievert2014ldavis}, showing topic overlap and a dynamic visual of the salient terms for each topic (Figure \ref{fig:topic-pyldaviz}). We use the same visual for the BERTopic topics even though it does not use LDA.

% \vfill

\section{Case Study 1: COVID-19}

The base of \systemname{} was originally designed for the Expressive Interviewing system as a response to the COVID-19 pandemic~\cite{welch-etal-2020-expressive}. The system uses a combination of techniques from Expressive Writing~\cite{pennebaker1986confronting} and Motivational Interviewing~\cite{miller2012motivational}. Expressive writing is a reflective writing technique shown to improve mental and physical health~\cite{frattaroli2006experimental}. Motivational interviewing is a counseling technique designed to help people change their behavior by eliciting one's motivation for change and reflective listening. 

\mysubsection{Objective.} Our original system aimed to reduce stress related to the pandemic.
More recently, ~\citet{stewart2023expressive}
used \systemname{} to perform a follow-up study to examine behavior change associated with the use of our system to understand how short- and long-term effects varied based on writing style and demographic factors.
Participants were asked about a variety of behaviors, such as how frequently they thought and talked about COVID or went out in public, and used these to determine if long-term behaviors had changed when they returned two weeks later.

\mysubsection{Interview Configuration.}
This interview consisted of four main questions. 
The questions centered around (1) the major issues in your life, (2) something you look forward to, (3) advice you would give to others, and (4) something you are grateful for (see Appendix~\ref{sec:appendix-interview-questions}, for the full questions). The interview contains reflections related to positive and negative emotions, pronoun usage, words related to order, and LIWC~\cite{pennebaker2015development} lexicon words related to money, health, home, and work.

Before the interview, participants were asked about their overall life satisfaction. After the interview, they were asked how personal and meaningful the interaction was. At both times, they were asked about their current stress level. All questions use a 7-point Likert scale. In a behavioral study, they were asked the additional behavior questions separately from our system, though these questions could easily be integrated as additional survey questions. The original study also asked users to compare our system to Woebot~\cite{fitzpatrick2017delivering}, a widely-used conversational mental help app.

\mysubsection{Participants.}
The original study recruited 174 participants through social media and announcements through our university.
For the follow-up study on behavior change, 200 participants were recruited from Prolific.co to use the system and return two weeks later to use the system again.

\mysubsection{Findings.}
In the original study, participants' self-reported stress levels were assessed before and after the interview, with $>3$ on a 7-point Likert scale being \textit{high stress}.
When comparing \systemname{} to Woebot, \systemname{} resulted in a greater reduction in stress (from 91\% to 64\%, a 9\% absolute decrease over Woebot). When asked to compare the two systems, users found \systemname{} to be easier to use and, overall, more useful~\cite{welch-etal-2020-expressive}.

When looking at the difference in participants returning two weeks later, writing with more lexical diversity was found to be correlated with an increase in social activity. Anxiety words in writing were correlated with stress reduction in the short term, and positive words with a meaningful experience. However, the short-term benefits to mental health did not translate to the long-term.
%length of interactions?

\section{Case Study 2: Neurotechnology - Brain Organoids}

Another study using \systemname{} focused on human brain organoids (HBO),  as an emerging technology in neuroscience research. Organoids, generally speaking, are miniature models of organs grown in a lab from stem cells that mimic the structure and function of actual organs and thus are a valuable tool for studying their development and health \cite{benito2020brain}.

\mysubsection{Objectives.}
The study aimed to replicate the findings from a previous survey on public attitudes toward organoid research~\cite{bollinger2021patients} and toward xenotransplanted chimeras, an organism composed of cells derived from different species that are though to raise ethical concerns for some individuals \cite{boyd2023scientific}. This work primarily surveyed the public about their moral attitudes toward brain organoids and/or chimeras.

\mysubsection{Participants.}
Students were recruited through the University of Marburg via an internal mailing list. In total, 39 participants used the system. Only five had heard of HBOs before using our system.

\mysubsection{Interview Configuration.}
The participants were asked eight main questions, consisting of (1) their initial thoughts about HBOs, (2) justification for their feelings, (3) conflict with moral beliefs, (4) transplanting human brain cells into other animals
, (5) impact on disease treatment, (6) views on suffering, (7) conflicted views, and (8) consciousness. Reflections related to sentiment and the LIWC categories of money and religion were also considered.

Before the interview, participants were asked if they had heard of HBOs before (yes/no) and if they were in favor of their use (this and subsequent questions presented on 7-point Likert scale). After the interview, they were asked how conscious HBOs would have to be before they are morally problematic, how meaningful and personal the interaction was, and again if they are in favor of their use.

\mysubsection{Findings.}
A manual analysis of the conversation topics reveals a large number of similarities with \citet{evans2022public}, where people express similar themes related to research benefits, ethical concerns with respect to consciousness, and cautious optimism. They similarly relate to the technology through science fiction references. They express unease with chimeric research, and religion plays a minor role in shaping opinions. 
The level of agreement in our study for HBOs and chimeric research were 84\% and 62\%, respectively, whereas previous work found these levels to be 90\% and 68\%. The findings are similar, though our sample size is small and derived from a different population (German versus USA). In contrast, the participants talked about the socioeconomic impact and how the technology might only be accessible to the wealthy. They also mentioned how the public must be able to give input and compared the technology to artificial intelligence.

\section{Discussion}

We demonstrated \systemname{}, a system that can be used by researchers to develop and administer research with a conversational interview style design. \systemname{} facilitates the collection of qualitative data that can provide deeper insight into respondents' perspectives and behaviors. Conducting these interviews through a conversational agent may also help users express perspectives more freely than they would to another person \cite{lucas2014s,gratch2014s}. By including pre- and post-interview survey options in the design, it offers a suitable framework to investigate the effects of the conversational interview. We demonstrated these utilities in two case studies that are contrasted by guiding conversations on familiar versus novel concepts. 

\mysubsection{Utility for chatbot intervention research.}
The COVID-19 study focused on guiding users to express their thoughts and feelings as a stress-reduction intervention using the expressive interviewing technique. \systemname{} allowed us to understand the effects of the technique on reducing stress. Many users who showed reduced stress after the interaction found the conversation meaningful. 
Also, users who found it more personal found it more meaningful as well.
This highlights opportunities to research how to personalize the experience~\cite{welch2021leveraging,abd2021perceptions}.

The Bioethics study aimed to gather ethical perspectives on a scientific topic that general populations may find novel, playing an educational role to facilitate this. Through the introductory text for the interview, studies can introduce participants to new concepts, or link users to external information or the FAQ section for more details. This allows the system to then capture first impressions of a new concept. The agent is not fluent enough to introduce these concepts in conversation, so researchers should take this into consideration.

\mysubsection{User Expectations.} Through interactions in both case studies, we find a small portion of users frustrated with the lack of natural interaction. We have three mechanisms for setting expectations; (1) the introductory text, (2) redirection to FAQ when a question is asked, and (3) encouraging participants to spend time writing their responses. We want them to view the system as a guide for them to write detailed responses rather than a quick back-and-forth chat.
Participants saw significant stress reduction and we were able to collect valuable input without a fluent agent, though the impact of fluency on the effectiveness of these methods deserves attention in future work.

\mysubsection{Considerations for Integrations with Generative AI.} 
\systemname{} is engineered as an end-to-end conversational system where the research admin has full control over the content the system produces. However, there are benefits that generative techniques could provide. 
For instance, we found that users of the COVID-19 system were more likely to find it meaningful if also personal, one direction could be to utilize LLMs to adapt to the user. Using LLMs to support personalization may be especially beneficial for systems like mental health chatbots~\cite{abd2021perceptions}.  

However, control over what the system generates is especially crucial for sensitive subjects like those explored in our case studies.
Generative models are unpredictable in nature and can produce harmful responses~\cite{bommasani2021opportunities}. 
How to mitigate or remove biases of these models remains an open issue~\cite{meade-etal-2022-empirical,blodgett-etal-2020-language}, leading researchers to suggest that they pose significant risks and should be carefully evaluated and monitored~\cite{laranjo2018conversational,dinan2021anticipating}, and that they may not be ready for many applications, especially those involving topics of a sensitive nature~\cite{lechner-etal-2023-challenges}.

Even if these issues were solved, generative models are still not to be desired for standardized conversational interactions. Standardized interviews, used for public opinion polling and academic research, influence decision-making in market research, corporate, and governmental organizations~\cite{gwartney2007telephone}. Often contrasted with surveys that allow for the collection of rich open-ended data~\cite{houtkoop2000interaction}, interview methodologies are more exploratory -- they vary in their level of standardization, with some allowing for more open-ended exploration~\cite{priede2011comparing}. This level of control over question standardization and exploration is not possible with language models. One of the salient problems with open-domain conversational agents is the lack of consistency in the generated output, often involving contradictions~\cite{roller2020open}.

\mysubsection{Extensibility.} 
Our open-source framework can be extended to pursue integration with other models, such as LLMs, that could modify output styles or add flexibility to conversation flow.
This feature would be optional and situation-specific as these components are often computationally or monetarily expensive~\cite{zhao2023survey}. Researchers could also extend it to offer human-in-the-loop features, enabling the use of LLMs to make the system more flexible while ensuring human oversight in sensitive scenarios.

\section{Conclusion}

We introduced \systemname{}, a framework for interview-style low-compute conversational agents that takes the form of a web interface with admin and participant facing pages. The interviews can be configured for new conversational topics through the administrative panel with no programming experience. By using a rule-based system, we avoid the drawbacks of generative models. 
By setting user expectations, \systemname{} can effectively help people reflect on their experiences and monitor behavior change (Case Study 1), and survey public opinion (Case Study 2). We release our code so that others can set up their own servers and extend our system.

\section*{Ethics statement}

The COVID-19 case studies were approved by the University of Michigan IRB. Case Study 2: Bioethics of Organoids was approved by the University of Marburg and University of Bonn IRBs.

\section*{Acknowledgements} 
We would like to thank Marlon May for helping with code compatibility, Ezzeddine Ben Hadj Yahya for his contributions to the code, and Tenzin Migmar for her testing and revisions to setup instructions.

\bibliography{rebib}

\appendix
\section{Interface Visuals}
\label{sec:appendix-interface}

This appendix contains a list of visuals of the interface for the major pages referenced and described in detail in \S\ref{sec:system-overview} and \S\ref{sec:administrative-panel}.

\begin{figure*}
    \centering
    \includegraphics[width=1.0\linewidth]{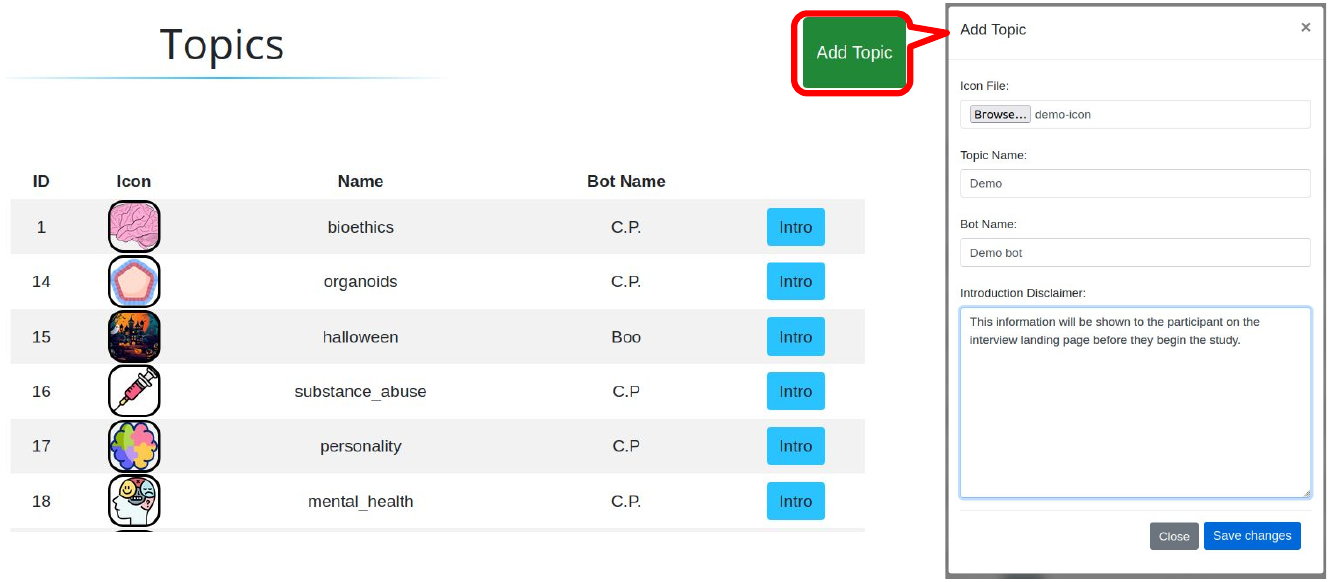}
    \caption{Topic page showing the list of topics and option to add a new topic, configuring the intro screen, name of the chatbot, name of the topic and icon representing the topic.}
    \label{fig:visual-topic-page}
\end{figure*}

\begin{figure*}
    \centering
    \includegraphics[width=1.0\linewidth]{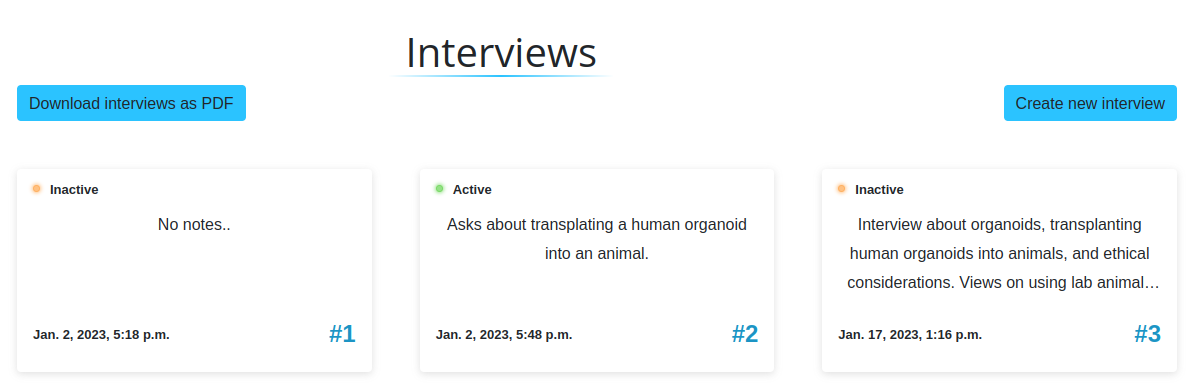}
    \caption{Interview overview page, showing all interviews configured for the topic as well as which one is active. The active interview will be the one used when participants start an interaction. Users can add new interviews or click on existing ones to view details.}
    \label{fig:visual-interview-overview}
\end{figure*}

\begin{figure*}
    \centering
    \includegraphics[width=1.0\linewidth]{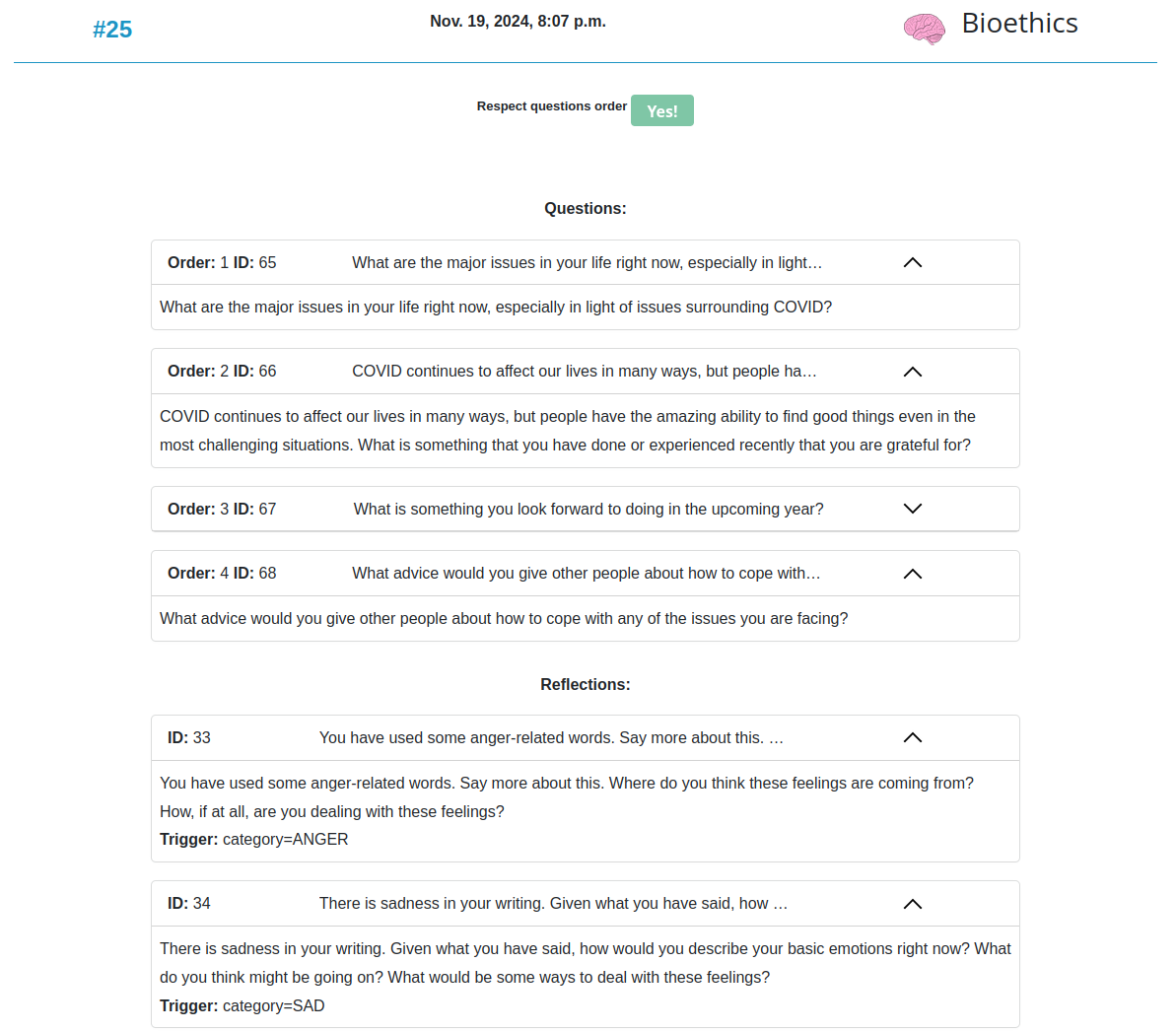}
    \caption{Interview details page showing the list of main questions that will always be asked and the list of reflections that can be triggered by different lexicons, sentiment, or conditions related to other reflections.}
    \label{fig:visual-interview-details}
\end{figure*}

\begin{figure*}
    \centering
    \includegraphics[width=1.0\linewidth]{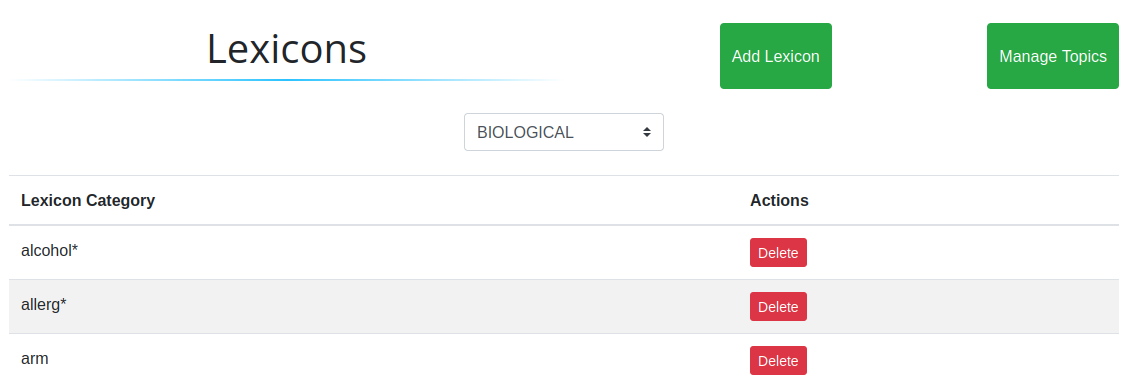}
    \caption{Lexicons page showing a dropdown with all lexicon categories. The list of words and word stems are shown for the selected category. Word stems are denoted with an asterisk at the end of the string. Users can add more terms or manage which topics use which lexicons.}
    \label{fig:visual-lexicons}
\end{figure*}

\begin{figure*}
    \centering
    \includegraphics[width=1.0\linewidth]{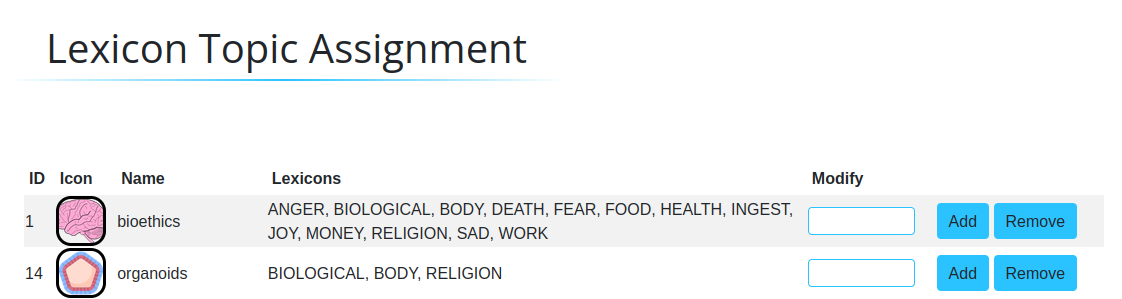}
    \caption{Lexicons topics page showing the list of topics and list of lexicon categories that are active for a given topic. Users can add or remove categories from each topic, affecting which categories are accessible as reflection triggers and which will show on the summary page and dashboard.}
    \label{fig:visual-lexicon-topics}
\end{figure*}

\begin{figure*}
    \centering
    \includegraphics[width=1.0\linewidth]{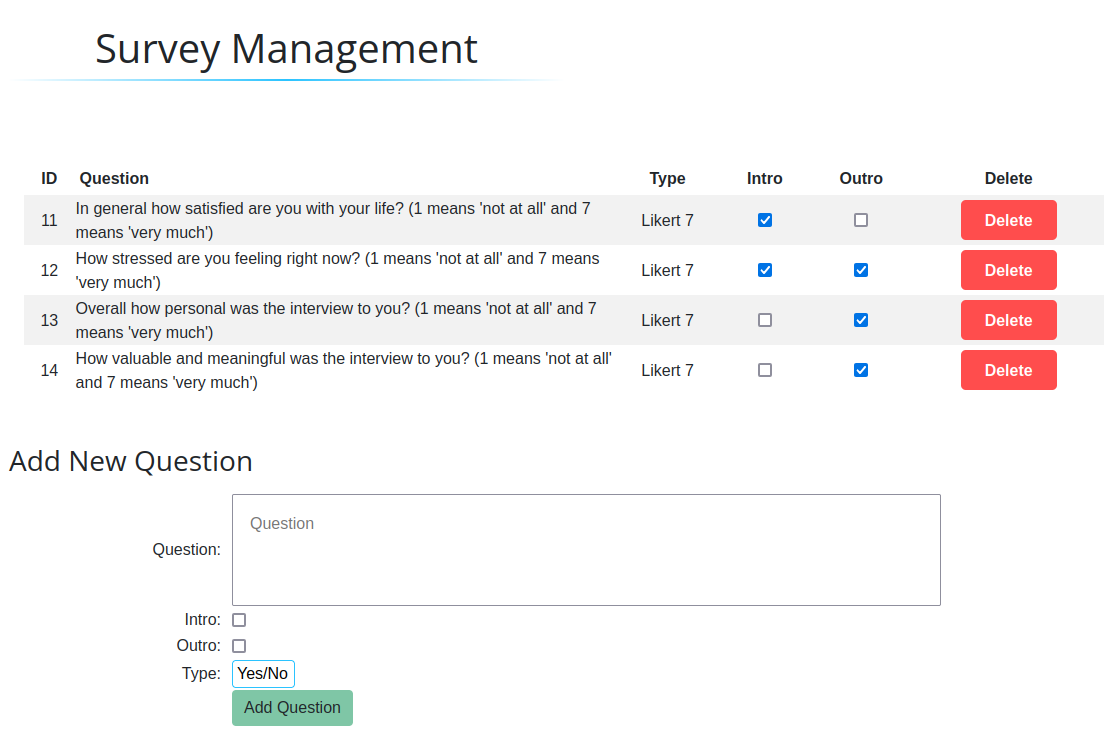}
    \caption{Survey management page showing the list of questions for the given conversational topic. Each question can be a yes/no or Likert scale question. Questions can be added, deleted, or toggled on/off for the intro and outro.}
    \label{fig:visual-survey}
\end{figure*}

\begin{figure*}
    \centering
    \includegraphics[width=1.0\linewidth]{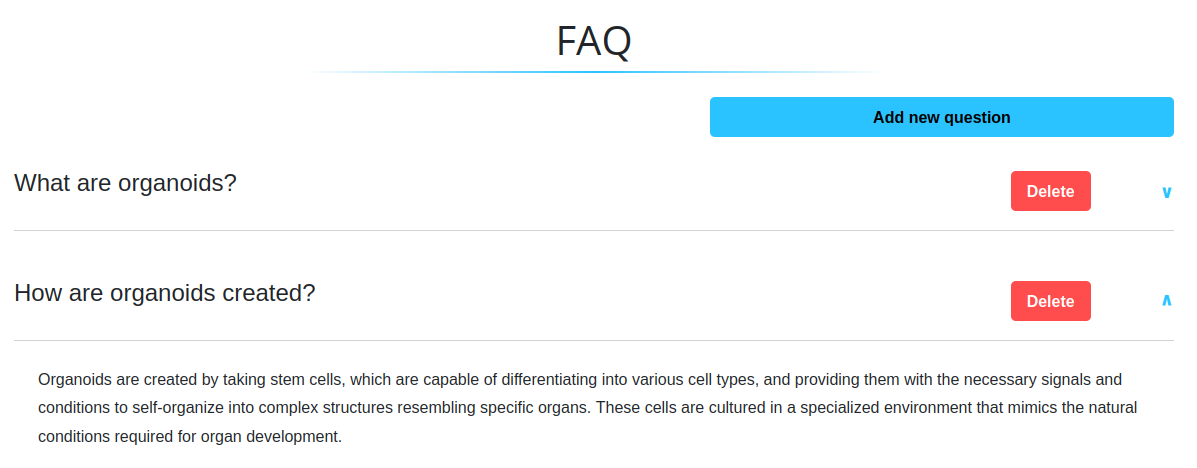}
    \caption{FAQs page showing frequently asked questions for a given topic. Questions can be added or deleted.}
    \label{fig:visual-faqs}
\end{figure*}

\begin{figure*}
    \centering
    \includegraphics[width=1.0\linewidth]{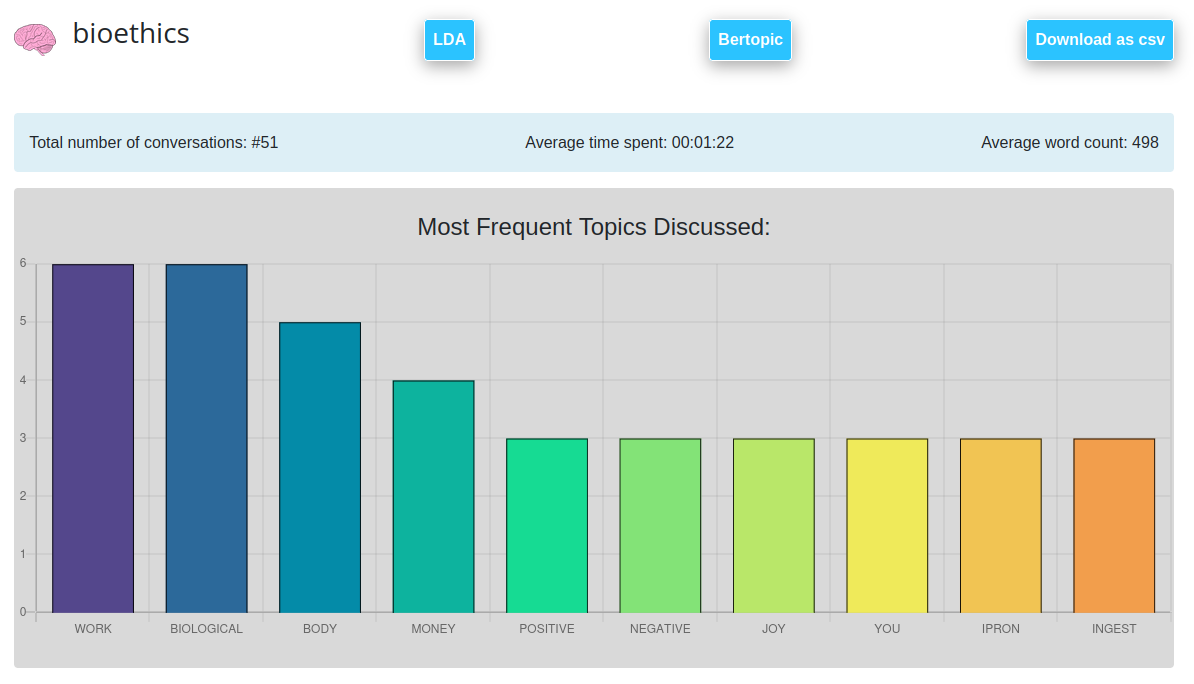}
    \caption{Top of the dashboard page showing a bar plot of the most frequently discussed topics, access to topic modeling, aggregate data download, total conversations, average time spent, and average word count, the latter two of which redirect to distribution plots.}
    \label{fig:visual-dashboard-top}
\end{figure*}

\begin{figure*}
    \centering
    \includegraphics[width=1.0\linewidth]{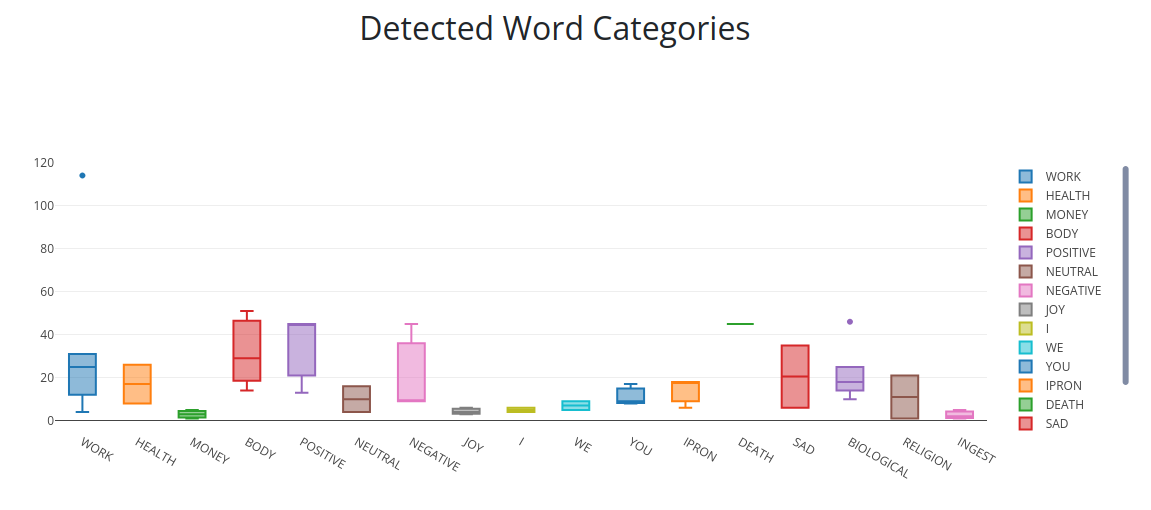}
    \caption{Dashboard plot of word category distributions. This shows how often each category tends to appear in a given interview for that topic.}
    \label{fig:visual-dashboard-word-categories}
\end{figure*}

\begin{figure*}
    \centering
    \includegraphics[width=1.0\linewidth]{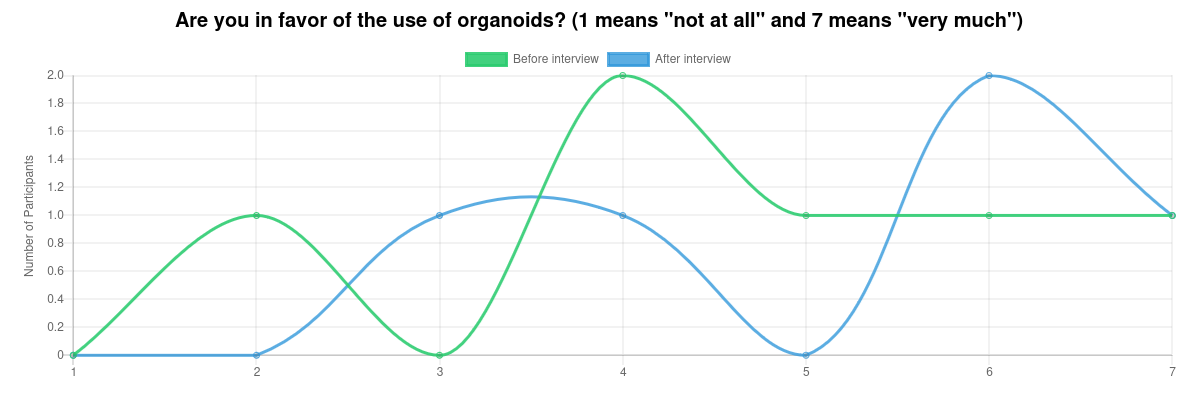}
    \caption{Dashboard plot generated for a Likert survey question. Each survey question has a generated plot with an entry for before and/or after the interview.}
    \label{fig:visual-dashboard-likert}
\end{figure*}

\begin{figure*}
    \centering
    \includegraphics[width=1.0\linewidth]{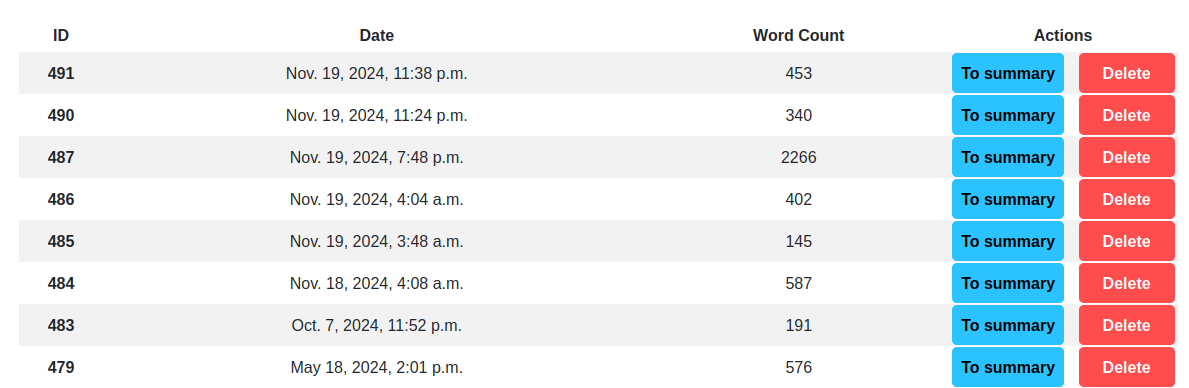}
    \caption{Dashboard list of conversation summaries. Each redirects to a summary page showing individual interviews.}
    \label{fig:visual-dashboard-list-summary}
\end{figure*}

\begin{figure*}
    \centering
    \includegraphics[width=1.0\linewidth]{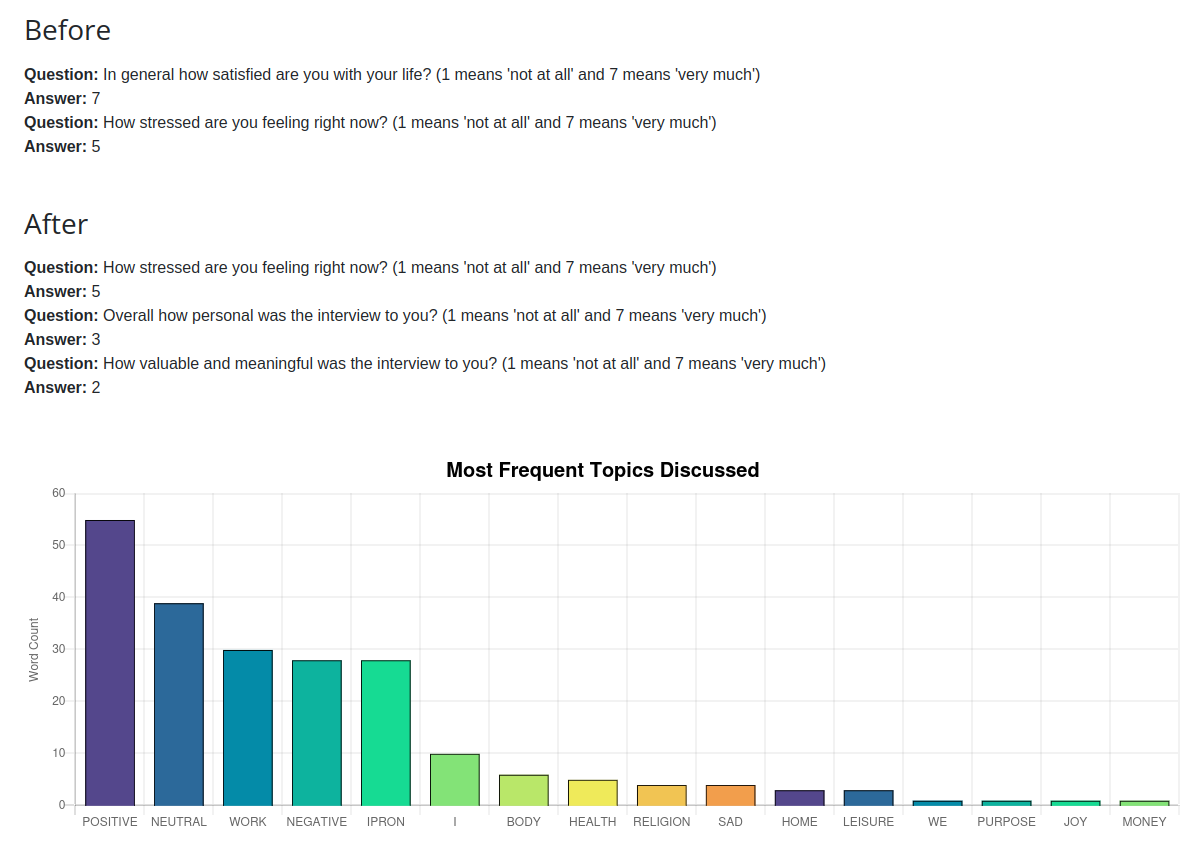}
    \caption{Summary page showing the survey answers and distribution of the frequency of each word category for that interview.}
    \label{fig:visual-summary}
\end{figure*}

\begin{figure*}
    \centering
    \includegraphics[width=1.0\linewidth]{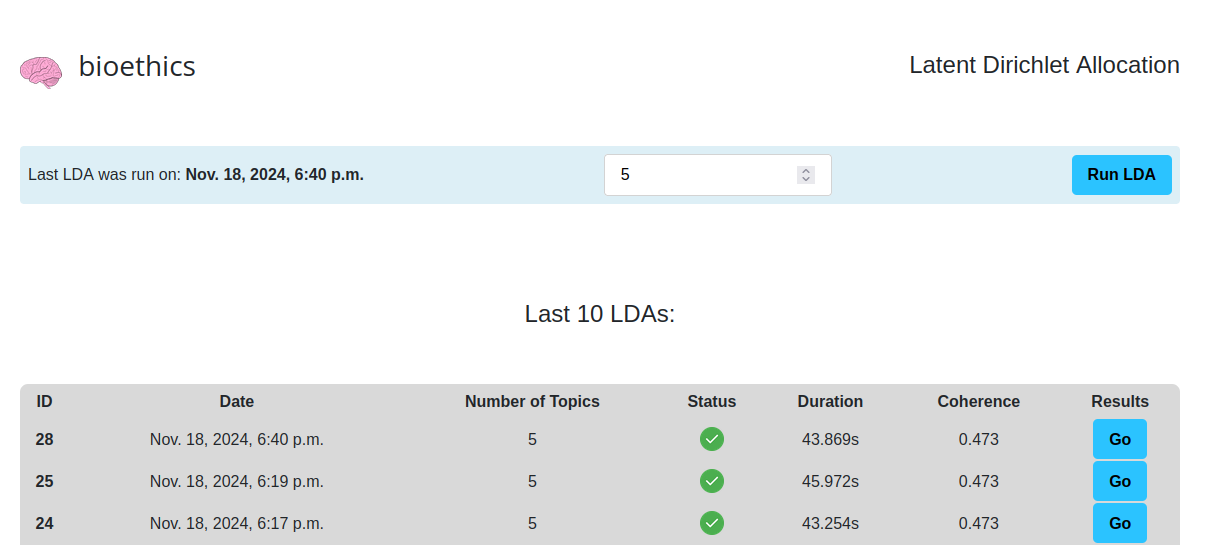}
    \caption{The topic overview page showing the history of LDA runs, the number of topics, duration of the run, and topic coherence.}
    \label{fig:topic-overview}
\end{figure*}

\begin{figure*}
    \centering
    \includegraphics[width=1.0\linewidth]{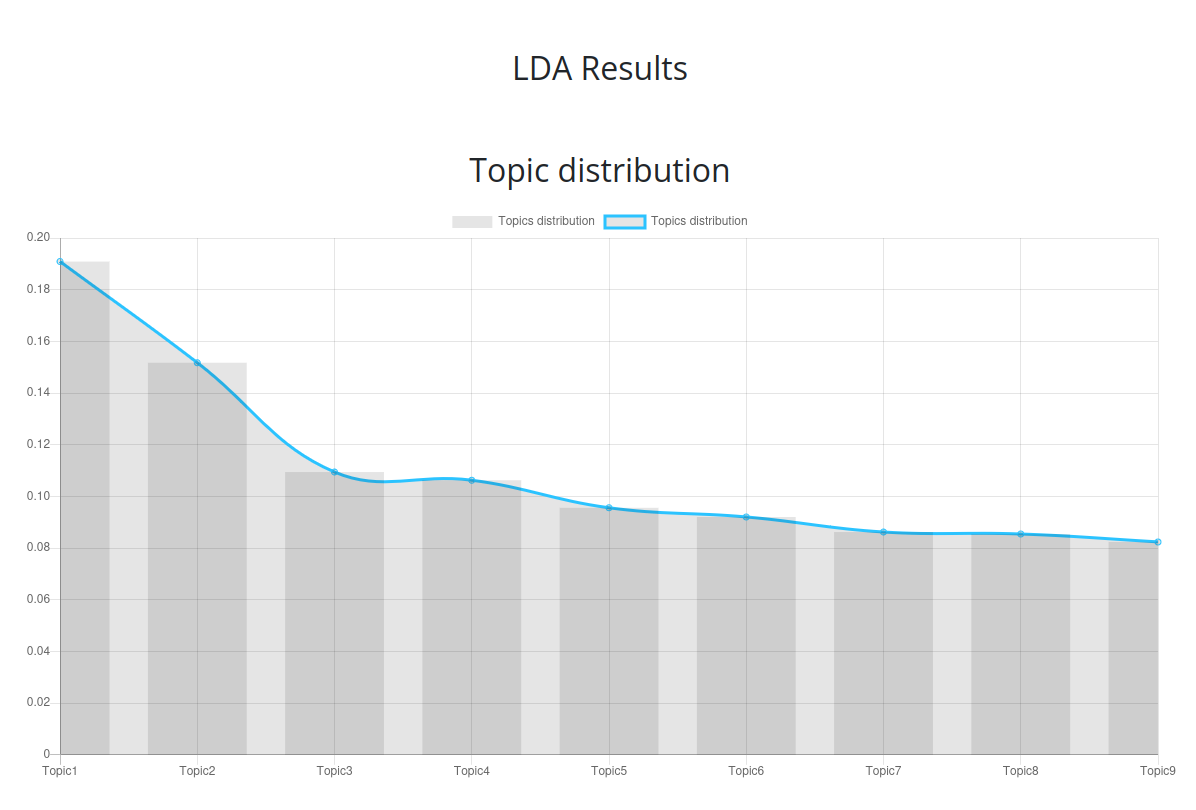}
    \includegraphics[width=1.0\linewidth]{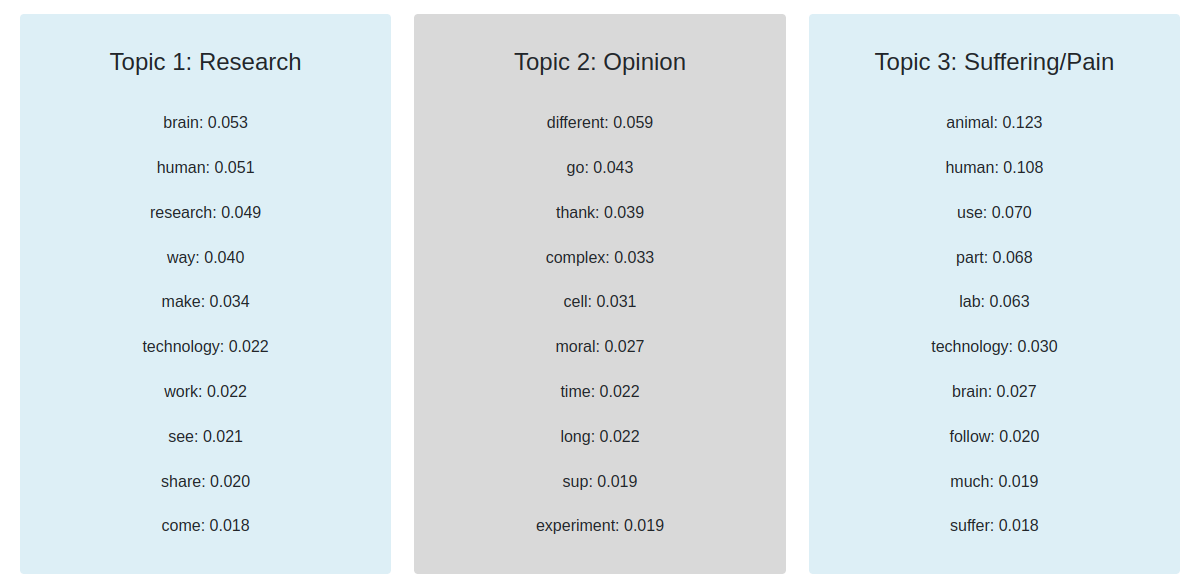}
    \caption{Top of the topic modeling results page showing the distribution of topics across all documents and top words in each topic.}
    \label{fig:topic-visual-page}
\end{figure*}

\begin{figure*}
    \centering
    \includegraphics[width=1.0\linewidth]{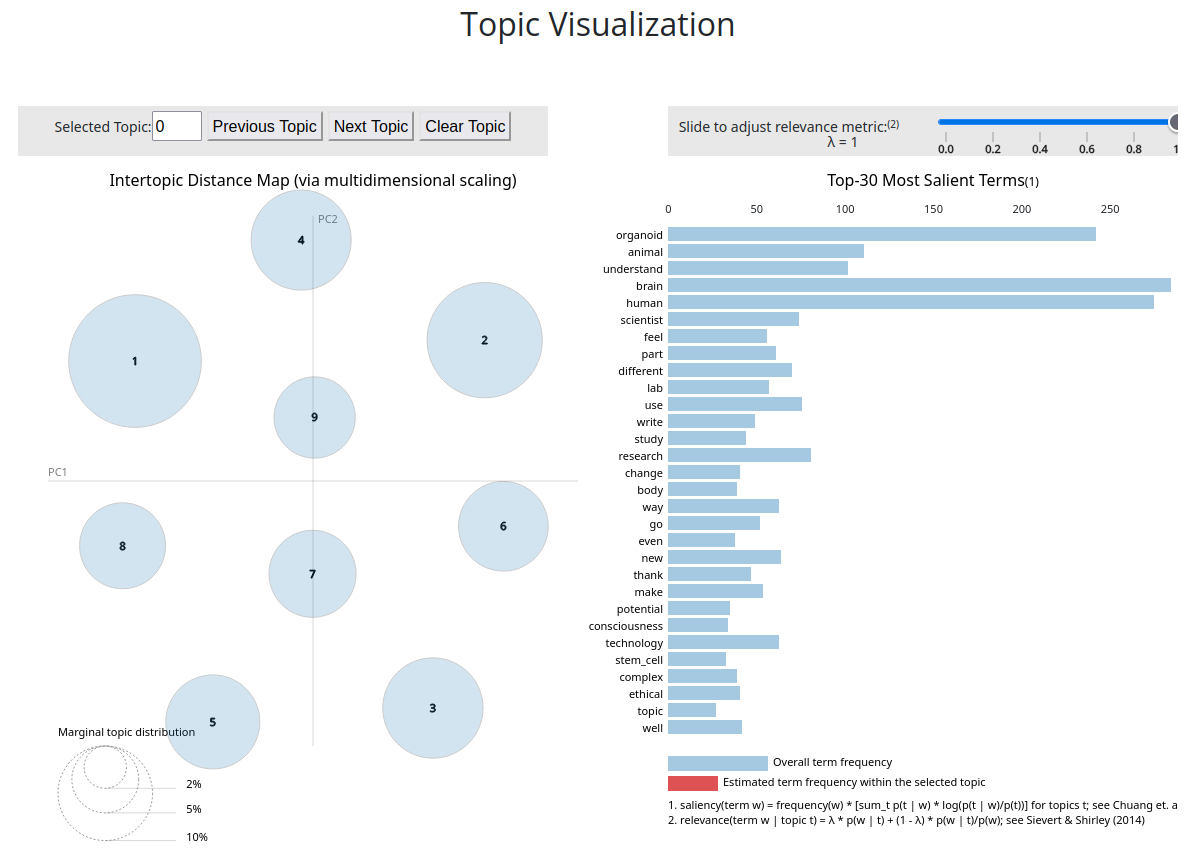}
    \caption{The topic visualization generated by pyLDAvis.}
    \label{fig:topic-pyldaviz}
\end{figure*}

\section{Interview Questions}
\label{sec:appendix-interview-questions}
This section contains the full interview questions mentioned in our case studies. For our original study on the COVID-19 pandemic our main questions were:

\begin{enumerate}
    \item What are the major issues in your life right now, especially in the light of the COVID outbreak?
    \item What do you most look forward to doing once the pandemic is over?
    \item What advice would you give other people about how to cope with any of the issues you are facing?
    \item The outbreak has been affecting everyone’s life, but people have the amazing ability to find good things even in the most challenging situations. What is something that you have done or experienced recently that you are grateful for?
\end{enumerate}

For the follow-up study on behavior change and short- and long-term effects, we used the same question about advice you would give to others but the other three questions had minor wording changes:

\begin{enumerate}
    \item What are the major issues in your life right now, especially in the light of issues surrounding COVID-19?
    \item What is something you look forward to doing in the upcoming year?
    \item COVID-19 continues to affect our lives in many ways, but people have the amazing ability to find good things even in the most challenging situations. What is something that you have done or experienced recently that you are grateful for?
\end{enumerate}

For our study on organoids, we used a larger set of questions:

\begin{enumerate}
    \item Organoids are a new tech for studying different parts of the human body in a petri dish. Scientists can use stem cells to grow different parts of the human brain. These human brain organoids can be connected to other organoids, or computers. What are your initial thoughts on human brain organoids?
    \item Can you share why you feel that way about brain organoids?
    \item Do brain organoids conflict with any of your core moral beliefs about what is right or wrong?
    \item How do you feel about the possibility of transplanting human tissue, specifically human brain cells, into other animals such as a mouse or monkey?
    \item Would your views on using lab animals with human brain cells change if the research was essential for developing more effective drugs for human disease?
    \item If you knew that the lab animals with human brain cells were not suffering and were treated better than normal lab animals, would your views on the use of these animals for research change?
    \item Are there any other aspects of human brain organoids that you feel conflicted about?
    \item Have you ever considered the potential for a brain organoid developing consciousness and emotions? If so, how do you feel about conducting experiments on such an organism, and do you think it could cause them suffering?
\end{enumerate}

\end{document}